# A Novel Approach for the Counting of Wood Logs Using cGANs and Image Processing Techniques

João V. C. Mazzochin [1], Giovani Bernardes Vitor [2], Gustavo Tiecker [3], Elioenai M. F. Diniz [4], Gilson A. Oliveira [1], Marcelo Trentin [1] and Érick O. Rodrigues [1,*]

[1] Graduate Program of Production and Systems Engineering, Universidade Tecnológica Federal do Paraná (UTFPR), Pato Branco 85503-390, PR, Brazil; joaomazzochin@alunos.utfpr.edu.br (J.V.C.M.); gilson@utfpr.edu.br (G.A.O.); marcelo@utfpr.edu.br (M.T.)

[2] Institute of Technological Sciences, Universidade Federal de Itajubá (UNIFEI), Itabira 35903-087, MG, Brazil; giovanibernardes@unifei.edu.br

[3] Business School, Universidade Federal do Paraná (UFPR), Curitiba 80060-000, PR, Brazil

[4] Graduate Program of Electrical and Computer Engineering, Universidade Tecnológica Federal do Paraná (UTFPR), Pato Branco 85503-390, PR, Brazil; elioenai@alunos.utfpr.edu.br

* Correspondence: erickr@id.uff.br

**Abstract:** This study tackles the challenge of precise wood log counting, where applications of the proposed methodology can span from automated approaches for materials management, surveillance, and safety science to wood traffic monitoring, wood volume estimation, and others. We introduce an approach leveraging Conditional Generative Adversarial Networks (cGANs) for *eucalyptus* log segmentation in images, incorporating specialized image processing techniques to handle noise and intersections, coupled with the Connected Components Algorithm for efficient counting. To support this research, we created and made publicly available a comprehensive database of 466 images containing approximately 13,048 *eucalyptus* logs, which served for both training and validation purposes. Our method demonstrated robust performance, achieving an average $\text{Accuracy}_{pixel}$ of 96.4% and $\text{Accuracy}_{logs}$ of 92.3%, with additional measures such as F1 scores ranging from 0.879 to 0.933 and IoU values between 0.784 and 0.875, further validating its effectiveness. The implementation proves to be efficient with an average processing time of 0.713 seconds per image on an NVIDIA T4 GPU, making it suitable for real-time applications. The practical implications of this method are significant for operational forestry, enabling more accurate inventory management, reducing human errors in manual counting, and optimizing resource allocation. Furthermore, the segmentation capabilities of the model provide a foundation for advanced applications such as *eucalyptus* stack volume estimation, contributing to a more comprehensive and refined analysis of forestry operations. The methodology's success in handling complex scenarios, including intersecting logs and varying environmental conditions, positions it as a valuable tool for practical applications across related industrial sectors.

## 1. Introduction

Applications that involve object segmentation and recognition in the literature are varied, from surveillance [1] and traffic analysis [2–4] to crowd management [5–7] and object counting [8–10]. In some of these cases, the counting process can be decisive, directly influencing the final outcome of a detailed analysis or serving as a means for system automation.

In an organizational industrial context—particularly in the production of physical artifacts—stock counting can be crucial for resource management. The counting of items in stock has a direct influence on the revenue and efficiency of an organization [11]. Technological innovations play a vital role in enhancing productivity and worker comfort. Solutions grounded in STEM (Science, Technology, Engineering, and Mathematics) significantly reduce inefficiencies and stress, showcasing the potential of data-driven approaches to optimize industrial processes based on statistical evidence [12]. In construction, precise log measurements are essential for estimating material requirements, ensuring structural stability, and minimizing waste [13]. For biomass energy production, accurate measurement is crucial for calculating energy yield, optimizing resource allocation, and enhancing the efficiency of production processes [14].

According to [15], at present, the wood_log measurement process relies on manual procedures conducted by workers using tape measures within the yard. However, this approach leads to inefficiencies, inaccuracies, safety challenges, and financial losses due to lack of automation. The manual nature of many processes is time-consuming and can lead to delays, causing a decrease in overall productivity. Additionally, relying on human judgment and precision introduces a higher risk of measurement errors, which can lead to incorrect data and subsequent decision making based on flawed information.

LiDAR-based methods have been explored as an alternative for automated log measurement, providing high accuracy and detail for 3D mapping and volume estimation [16]. However, such systems require significant investments in hardware and expertise, which may limit their widespread adoption in smaller-scale industries. Additionally, the reliance on specific equipment makes these methods less adaptable to different operational environments.

In contrast, our method offers a simpler and more accessible solution by directly processing 2D images instead of requiring 3D data acquisition and analysis. This approach reduces the complexity of implementation, eliminates the dependency on specialized hardware, and makes the system more adaptable to a wider range of operational contexts. By leveraging computer vision techniques, our solution is not only cost-effective but also scalable, addressing key limitations of existing automated systems.

Furthermore, inaccuracies in manual measurements can have significant financial consequences [11]. Incorrect measurements may result in overestimating or underestimating the quantity of materials, leading to inefficient resource allocation, sub-optimal inventory management, and potential financial losses. Inaccurate measurements can also affect pricing and billing processes, resulting in disputes and potential loss of revenue.

By transitioning from manual measurement to consistent automated systems, businesses can streamline their operations, ensure accurate data collection, promote worker safety, and optimize resource allocation. Addressing these challenges requires automated approaches, leveraging computer vision, machine learning, and image processing techniques.

In this work, we address the following research questions:

1. How can automated wood log counting systems be designed to overcome inefficiencies, inaccuracies, and safety challenges associated with manual methods?
2. How effective is the cGAN framework in addressing the specific segmentation and counting challenges presented by wood logs, a context often underexplored in the literature?
3. What are the limitations of combining deep neural networks and mathematical morphology for object segmentation and counting?

To answer these questions, we propose the following:

1. A novel dataset of wood logs captured by the authors of this work;

2. An approach to the automation of wood_log counting, which is still not yet properly explored in the literature;
3. A novel methodology from the algorithmic standpoint, where we combined existing techniques and approaches from machine learning and image processing to generate our final result.

In this sense, we use the Pix2Pix framework [17], which is an image-to-image translation framework not actually coined for object counting or image segmentation. No previous records were found regarding the utilization of Pix2Pix for object identification in images, although it has already been used for image segmentation in other image contexts [18]. Moreover, few studies address a similar context involving the identification and counting of wood logs using segmentation. Among these studies, only [10] provides sufficient data for a minimal (not yet fair) comparison, which also reinforces an adequate and further exploration of the topic in the literature.

In terms of technical aspects, our approach uses a combination of multiple techniques: first, (1) we use the Pix2Pix framework to segment the pixels using the generic problem description "rings of the wood vs. non-rings" pixel; later, (2) we employ mathematical morphology [19] to refine the output of the neural network; and third, (3) we use the Connected Components Algorithm [20] to output the actual amount of logs in the image.

## 2. Materials and Methods

One of the main benefits of putting efforts into acquiring a database is fine-tuning the parameters and the approach to deal specifically with a certain type of data, which can enhance the result of the model. Several works in the literature use photographs of types of wood that are not the object of study and deviate from the core application of this methodology, which is centered on a rural area of Brazil, the city of Pato Branco in the state of Parana. Due to that fact, we focus directly on the type of wood available to the industry in the region, the *eucalyptus* trees and their wood logs, which are regionally used to power engines, in construction and civil engineering, to create paper and furniture and even to heat homes during winter. Furthermore, we publicly provide the collected dataset for further comparison, replication, and usage in research.

The Brazilian Institute of Geography and Statistics (IBGE), in charge of producing and analyzing statistical data in Brazil, reported that the forestry sector achieved a total production of 158,283,790 $m^3$ of wood logs in the country. Of this quantity, 99,693,522 $m^3$ were used for the manufacture of paper and cellulose, while 58,590,268 $m^3$ were used for other purposes. It is important to highlight that around 69.70% of this amount of wood in logs comes from *eucalyptus*. This statistical information was the reason for creating a database that covers wood of this species [21].

Considering that it is unlikely to compile a dataset encompassing all types of wood for all types of applications, as this would most probably harm the trained model efficacy, we emphasize that the weights generated by the network trained in this study can serve as a solid starting point for transfer learning to other types of wood logs. This strategy not only enhances the scalability of the proposed model but also reduces computational costs and training time when adapting it to new conditions in corroborative works aligned with our methodology. Furthermore, we will make the trained model's weights publicly available along with the collected dataset, allowing other researchers to utilize and build upon them in their own studies.

The database used in the training was collected with images that mainly vary with regard to the following features:

- Light intensity and light-related characteristics;
- Capture angle and displacements of the logs;

- Distance from the object;
- Natural variation in the color of tree logs;
- Images containing extra details that present challenges during segmentation, such as leaf overlapping, branches, and barks.

The images were captured by two mobile devices with 12 megapixels (Xiaomi Pocophone F1, Beijing, China) and 13 megapixels (Samsung Galaxy J6+ - Seoul, South Korea), which outputted images with a size of 4032 $\times$ 4032 pixels, which were later downscaled to 256 $\times$ 256 for training, due to the limitation of several neural network architectures. The images contain *eucalyptus* wood logs in piles, the usual way in which these logs are stored. The database has a total of 466 images that were captured from 8:00 a.m. to 5:00 p.m. in order to accommodate different levels and incidences of illumination. The dataset, including the annotated ground truth and the trained weights of the Pix2Pix network, is publicly available on GitHub [22]. Below, we provide further details regarding the acquisition protocol.

- A total of 167 images were captured from a close frame, a narrower field of view, keeping the number of woods close to 25. The capturing of the images was carried out in favor of the sun or with the sun on the right side, that is, without great variation in lighting and beams of light. However, the presence of shadows is notable due to the variation in the depth of the wood logs due to the positioning of the sun. In these images, the main characteristics that present a challenge to the identification model are the variation in size, face coloration, difference in depth between wood logs, angle of capture, and overlapping of wood logs. Figure 1 presents an example of close-frame images displayed in the second row.

  Despite the relatively simple nature of these images, several factors introduce challenges to the segmentation and identification of individual wood logs. The overlap between logs often obscures portions of their structure, making it difficult to clearly distinguish their shapes. Additionally, the variation in heartwood color can lead to visual confusion during the object segmentation process. While not predominant, the intricate texture of the bark, coupled with differences in size and shape, adds further complexity to automatic detection. Irregular lighting within the stack, resulting in alternating dark and bright areas, poses additional obstacles. Lastly, debris and foliage present between the logs introduce extraneous elements that can negatively impact detection accuracy.
- A total of 317 images were obtained from an open frame, a wider field of view. Some samples contain hundreds of logs, while others have intense beams of light and a significant presence of overlapping leaves and branches. Due to these specific characteristics, those images were not utilized for statistical analysis in this study, but they hold potential for further exploration in future research.

For the wider-frame images, it becomes evident how challenging it is to identify individual wood logs. Not only do the aforementioned issues become more pronounced, but the lighting within the stack, creating variations in brightness, further complicates the identification process. The existence of debris and foliage scattered among the wood logs introduces additional factors that may disrupt the accuracy of detection. Lastly, the combination of the camera's focal length and sun-induced glare has the potential to distort the proportions and shapes of the wood logs, directly impacting the detection system's ability to accurately interpret the scene.

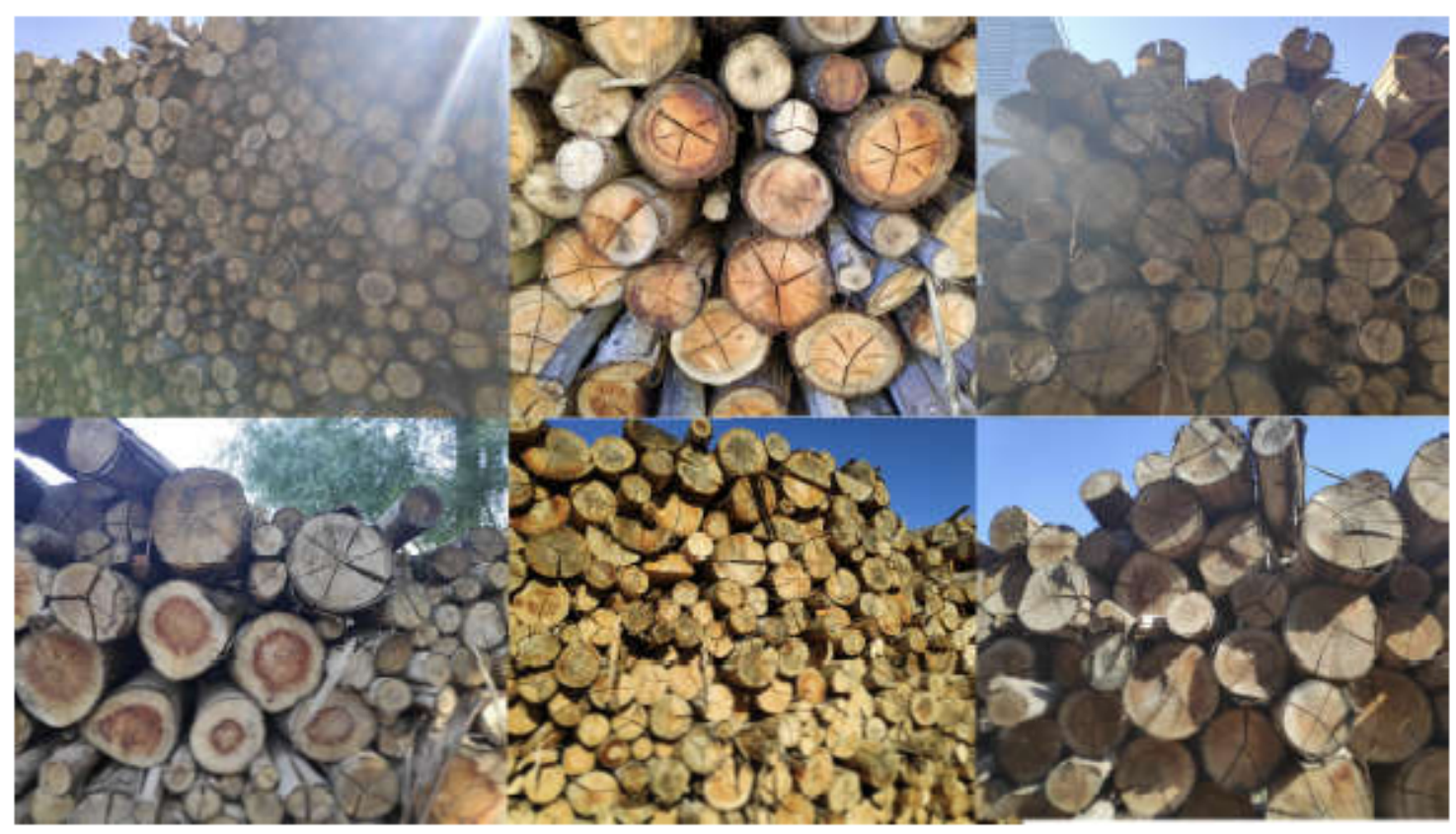

**Figure 1.** Samples of the images captured from a closer and wider frame.

Figure 1 shows a fragment of the database used in this work. In this case, only close-frame images were used. Close images allow for a more focused and precise analysis of the object of interest without the presence of distracting elements in the background (branches, leafs, beams, etc.).

This research focused on a more controlled scenario, where the user that is taking photos of the piles is obliged to follow certain guidelines in order to sustain the accuracy reported in this work. Our approach is capable of working on different scenarios that consider different angles, illumination, occlusion, and positioning, but the user taking the photos is expected to produce images that, in a sense, look similar to what we have in the dataset that was used for training. Some of the issues that arise are highlighted below.

- Object positioning: the precise placement of the object within the frame can vary in terms of scale, orientation, and overall positioning. The angles of acquisition also play a major role at this point, which can be difficult to overcome depending on the chosen technique.
- Occlusion: we have to deal with different degrees of occlusion. It occurs when parts of the object are hidden or obscured by overlapping elements within the frame, such as other objects or other wood logs. Some occlusions are also generated by the angle of acquisition of the images.
- Light conditions: variations in daylight and sky conditions, including changes in brightness, shadow presence, and overall illumination, can significantly affect the object appearance in the photo. These conditions also influence contrast and color temperature throughout the day, contributing to potential image artifacts such as lens flares.

Furthermore, image quality is also a very important factor. Variations in image resolution, sharpness, and overall clarity can have implications for algorithmic performance degradation and the reliability of the obtained results.

The use of machine learning and image processing to automate classification tasks is aimed at seamless automation, as manual execution often leads to inconsistencies among different experts [23].

The pixel-by-pixel segmentation process was carried out using the Pix2Pix framework, which operates based on the paradigm of instance (or object) segmentation. The diagram in Figure 2 provides a detailed visualization of the steps involved in training the model. The process begins with the manual segmentation of the original image to create the ground truth. The pre-processing stage, illustrated in the central section of the diagram, includes three distinct methods: gradient operations in gray scale, gradient operations in color, and erosion. Each method was tested individually to assess its impact on the segmentation process. Among these methods, the erosion method demonstrated the highest success rate in the segmentation task, as indicated by the bold lines in the diagram. This configuration was identified as the most effective workflow for achieving optimal results when used as input for the Pix2Pix model.

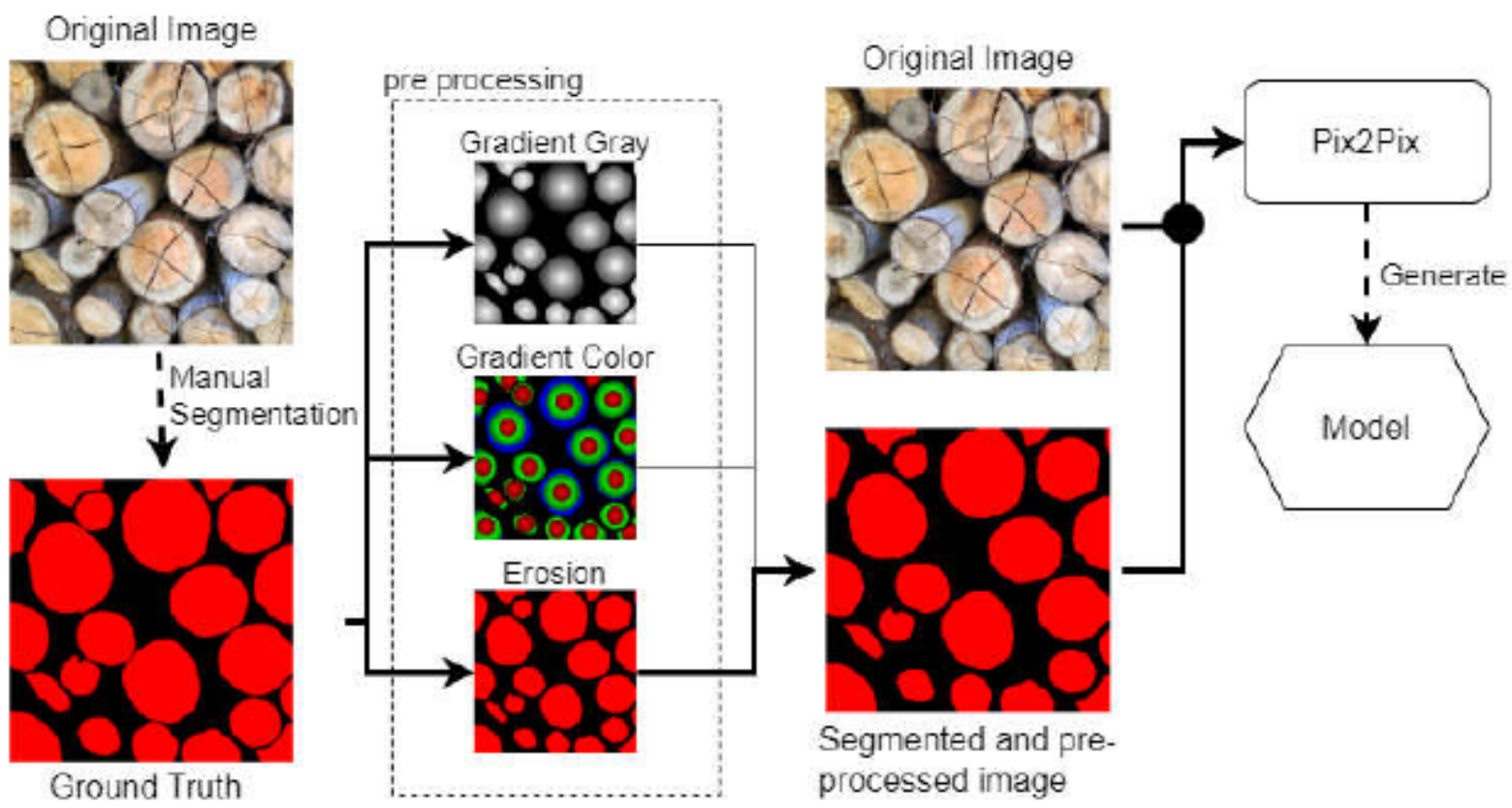

**Figure 2.** Overview of possible training steps for the proposed methodology for segmentation.

It is worth noting that both the original image and the pre-processed segmented image generated by each of the three pre-processing methods were used as inputs for the Pix2Pix model during training. Variations in the training parameters $input_{nc}$ and $output_{nc}$ were explored to control the number of color channels in the input and output images. These adjustments allowed for flexibility in adapting the dataset, such as switching between grayscale (single-channel) and RGB (three-channel) images.

To evaluate the effectiveness of the proposed methodology, we conducted comparative experiments with three neural network architectures: a Vanilla Convolutional Autoencoder, a Vanilla U-Net, and the Pix2Pix architecture. These architectures were selected for their varying complexities and learning capacities, allowing us to explore the relationship between the number of parameters and model generalization in the context of wood log segmentation.

The models were tested using the same images of the dataset created for this study, consisting of images with varying lighting conditions, capture angles, and overlapping log types. During preliminary tests, both the Autoencoder and U-Net exhibited clear signs of overfitting when configured with more than 20,000 parameters. This behavior was evidenced by an increasing gap between training and validation losses and poor generalization to new input conditions. To address this issue, we reduced the number of filters in both architectures, resulting in simplified versions that still showed limitations in their final performance.

In contrast, the Pix2Pix architecture, used as the core of our methodology, efficiently handled more than 57 million parameters without exhibiting signs of overfitting. The robustness of this approach can be attributed to the use of conditional adversarial learning, where the discriminator regulates the training of the generator, and the higher diversity of the dataset.

The detailed results of this comparison, including quantitative and qualitative analyses, are presented in Section 3.

Continuing with the pre-processing techniques. As shown in Figure 2, we evaluated three types of pre-processing for the ground truth (manually segmented wood logs) in order to confirm whether any of them improve the output segmentation of the Pix2Pix network.

The three approaches consisted of (1) replacing each log with a circular gradient that is brighter in the center and darker at the edges. We created this experiment to check whether the Pix2Pix framework would be able to differentiate the central region of the wood (heartwood) from the rest. The transformation from the binary ground truth to the gradient ground truth can be seen in Figure 2.

The (2) second approach was an experiment with a gradient varying in color instead of just gray scale. This hypothesis stems from the fact that Pix2Pix is a color-oriented framework. We tried both gray-level input and colored input just to check whether we can obtain an improvement in any case. In theory, gray level equals less information to train, but in the end, the important factor is how the framework uses the information in both cases.

The (3) third and final pre-processing approach consists of applying a morphological erosion in the ground truth. This assumption relates to the bark of the tree; we aimed to reduce its influence in the prediction once again aiming towards better segmentation results.

After the segmentation phase, we still need to count the objects in the output image, as this is not a direct output of the Pix2Pix network. We evaluated the following approaches for the counting: Circular Hough Transform, Connected Components, and Morphological Reconstruction. A visual representation of our counting methodology is depicted in Figure 3.

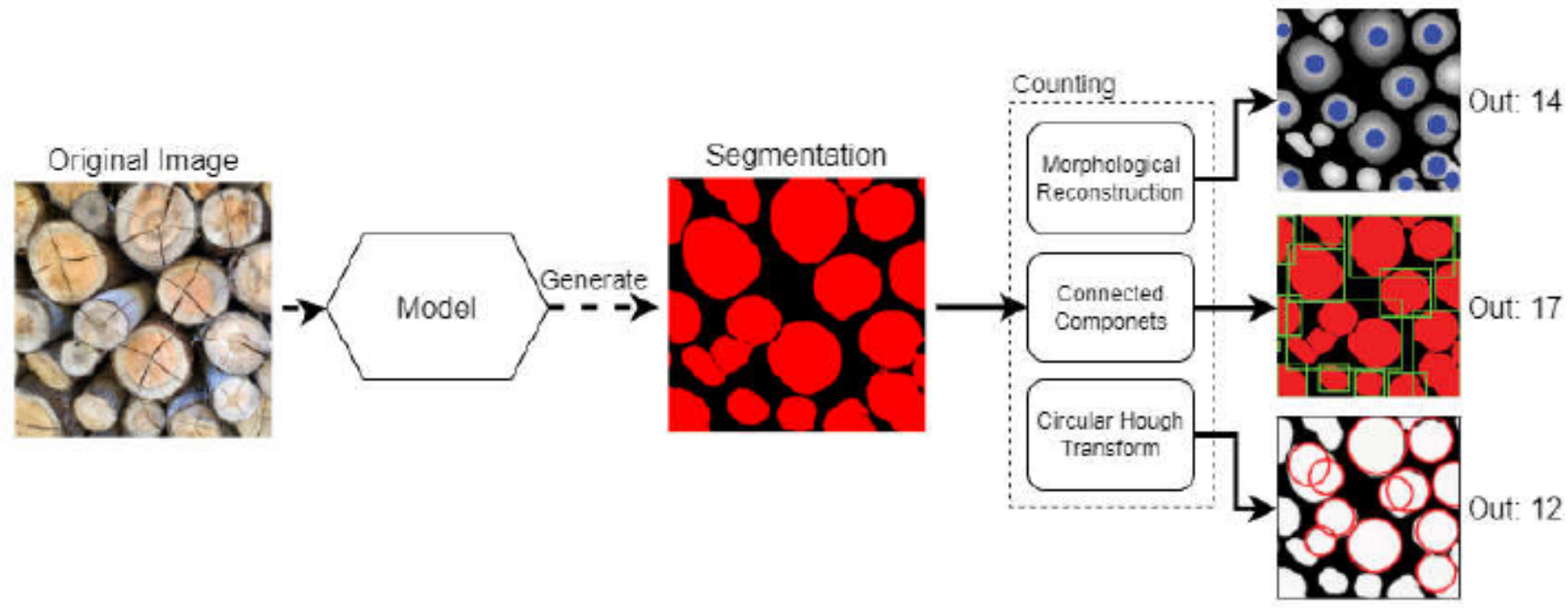

**Figure 3.** Counting methodology overview.

Initially, when a novel image, not part of the training dataset, is fed into the model, the Pix2Pix model is required to create the corresponding segmentation. Subsequently, the image is subjected to the counting algorithms, which in turn can output another image, highlighting the identified wood logs.

The ground truth segmentation was subjectively conducted, adhering to guidelines gathered during the research, including not overlapping wood_log segmentation; adhering to the bounding edge of wood logs; and avoiding the segmentation of adversities such as leaves, branches, and tree barks.

In terms of intersections, the initial generations of the Pix2Pix predictive model, we noticed that due to the downscale process carried out by the framework, the images that previously clearly depicted the separation of wood logs with at least 1 pixel of "valleys" between different woods were now merging the segmented logs in the ground truth during training.

This was a major issue since the model was capable of producing fairly realistic segmentation, but they usually resulted also in numerous intersections. As we are counting the objects from a post-processing standpoint, this posed a challenge.

To solve this problem, we assumed that applying a morphological erosion [19] as a pre-processing step would get rid of the intersections and improve the results. While conducting the research, we considered applying erosion to the images generated by the model. However, we identified that erosion not only would eliminate intersections and noise but also could remove segmented wood logs with a small diameter. Based on these considerations, we determined that 15 iterations of erosion during pre-processing were the most effective choice for removing intersections and noise in our dataset without penalizing small segmented wood logs.

To further analyze the impact of erosion, we conducted experiments applying erosion with values ranging from 5 to 50 pixels in steps of 5. The results are summarized in Figure 4, evaluating the performance of the models under these conditions.

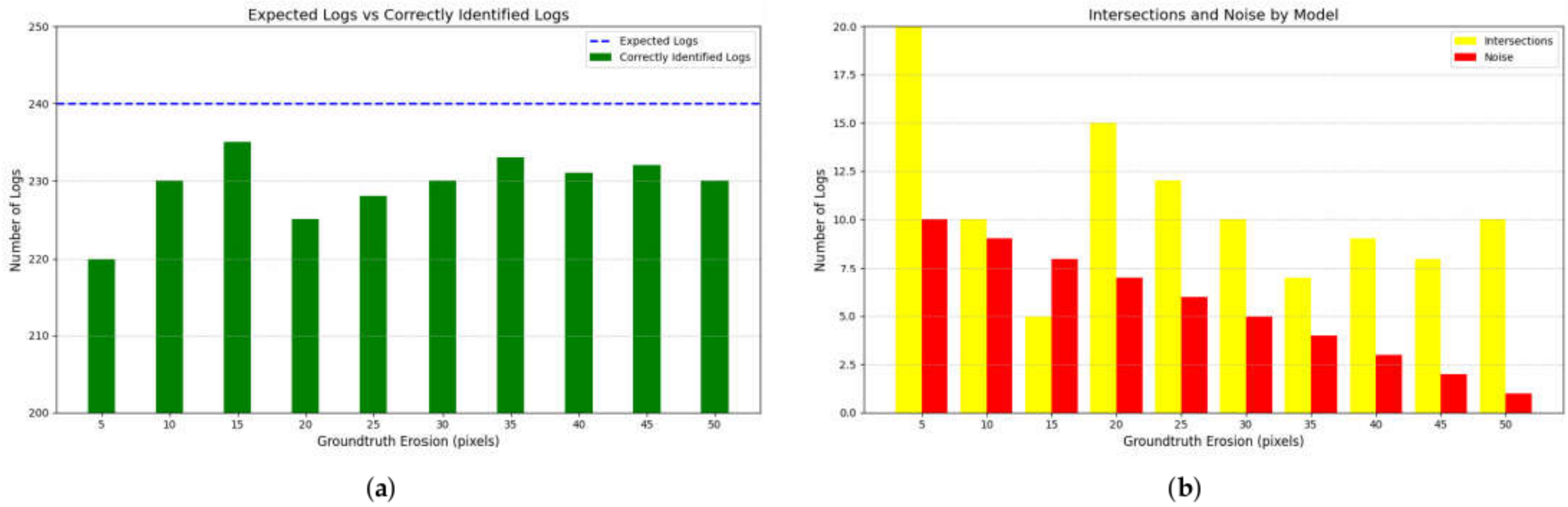

**Figure 4.** Performance analysis of models under varying levels of ground truth erosion: (**a**) compares the expected logs with the correctly identified logs, while (**b**) focuses on intersections and noise at different erosion levels.

These additional experiments provide a comprehensive understanding of how varying levels of erosion influence the balance between removing noise and retaining smaller wood logs, offering insights into optimizing this pre-processing step for different datasets.

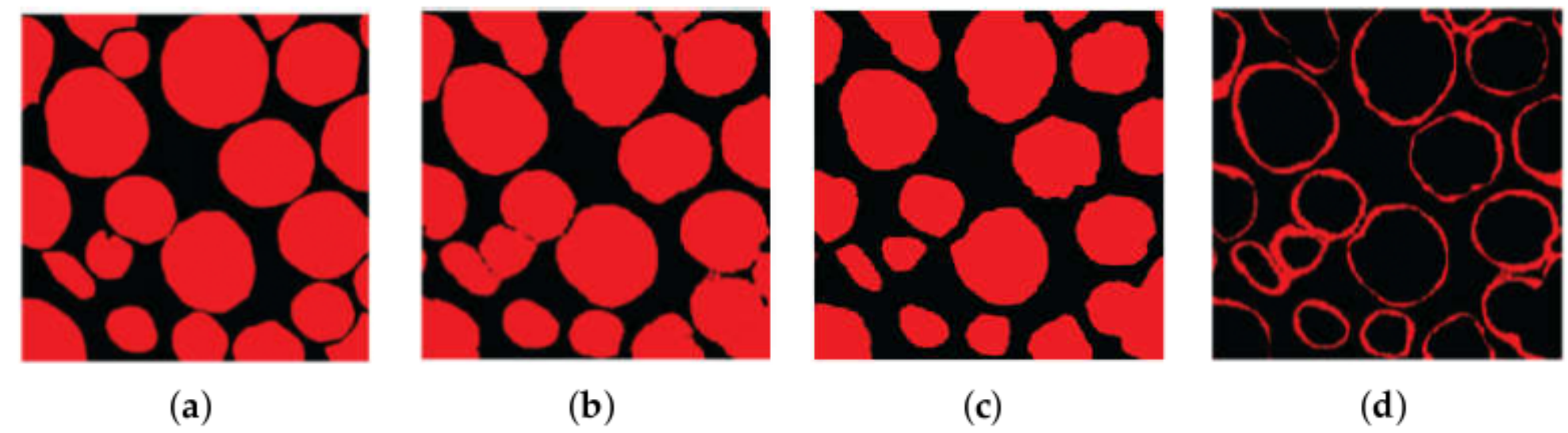

**Figure 5.** Comparison of trainings with and without erosion as a pre-processing step: (**a**) ground truth segmentation; (**b**) generation without pre-processing; (**c**) generation with pre-processing: erosion; and (**d**) subtraction of (**b**,**c**).

The improvement is evident when highlighting the division between groupings. Even though some intersections persist, it is clear that the model has improved its ability to avoid overlaps and noise. This evolution is notable when examining the more defined separation between the clusters in Figure 5c.

The gradient idea, which was previously mentioned, was used to check whether we could better identify the centroids of each circle representing the wood_log, maybe leading to new ideas for counting. In summary, we trained 3 different models using (1) the regular red color for clusters, (2) a gradient red that is brighter towards the center of the circle and has a limited diameter, and (3) the same gradient but in gray color and considering the entirety of the circle. The second approach was tested because the irregularities in the diameters lead to some difficulties in the counting phase due to the intersections. One of the ways to remove the intersections is applying erosion, but it also gets rid of the smaller circles in the segmentation output. Figure 6 depicts these 3 discussed variations.

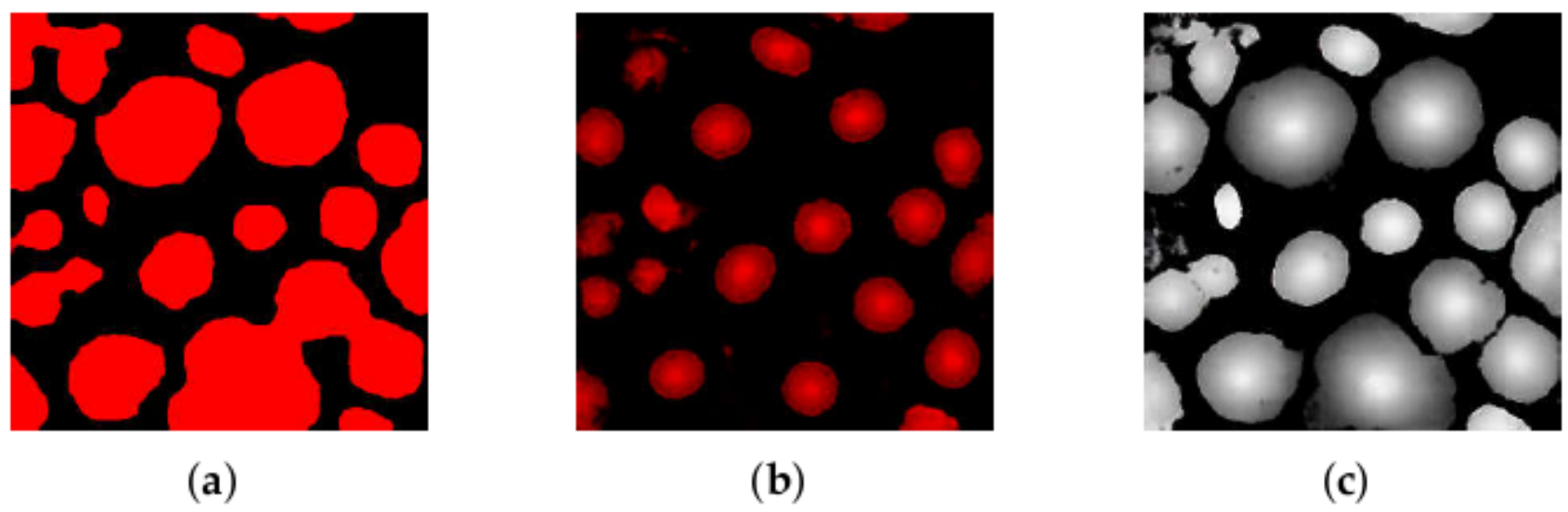

**Figure 6.** Examples of the three test cases used for segmentation evaluation during this study: (**a**) plain red masks without gradients, (**b**) red gradient with a fixed maximum diameter, and (**c**) grayscale gradient preserving the original object diameter.

The models were trained using 1328 logs (2,970,501 pixels) from the training dataset, and an additional set of 448 logs (1,002,097 pixels) was reserved for testing purposes. The results of these tests are illustrated in Figure 7.

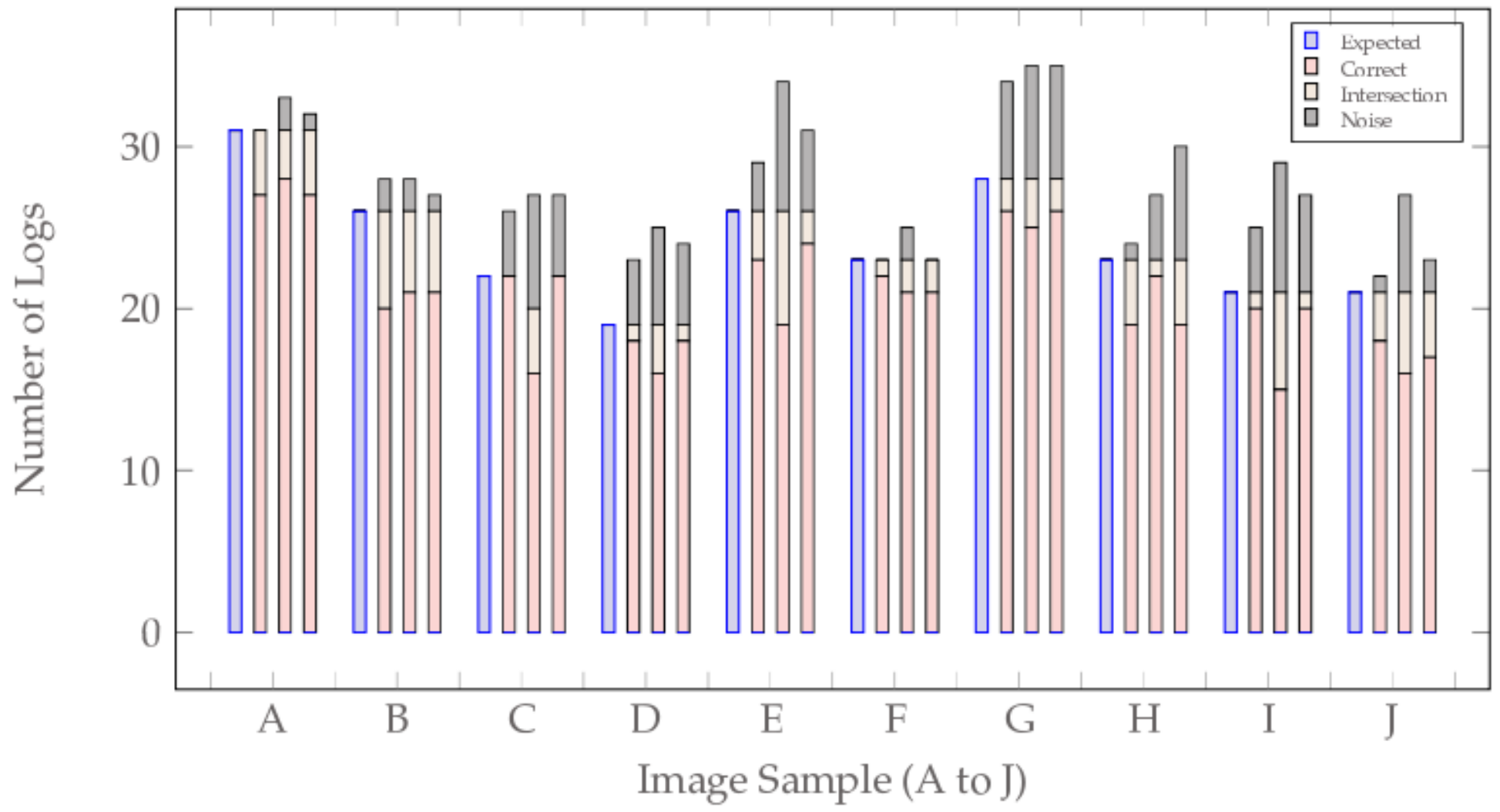

**Figure 7.** Model training on ground truths (1), (2), and (3). The letters A to J represent an image sample from the testing dataset.

Each letter on the X-axis corresponds to one image sample from the test dataset. The Y-axis denotes the number of logs identified. The blue lines represent the expected number of logs for each testing sample, while the stacked bars illustrate the results produced by 3 different models from training cases 1, 2, and 3 mentioned above.

The green bars represent wood logs correctly identified by the model, whereas the yellow part indicates logs that were identified but with intersections. At last, the red color represents the noise introduced by the model. This analysis is not a final result of the proposal and was used to guide our analysis if there is any need to pre-process the ground truth in order to improve the final segmentation results.

An analysis of the collected data reveals that the model with flat red segmentation achieved the highest performance, excelling in reducing intersections while minimizing noise generation. Figure 6 presents a sample of the segmentation outputs generated by models (1), (2), and (3), which provided the most effective results.

Still, the gradient gray remains a prospective solution, as articulated further in the manuscript (Section 5). However, for the scope of this study, the segmentation model was configured to generate outputs utilizing the plain red color. It provided nice results and is faster as it does not add any calculations and changes to the original ground truth.

Building upon the segmentation generated by the model, we adopt the identification by clusters. As previously mentioned, the main obstacle lies in the intersections between different groups, which directly impact their identification, as once intersected, these groups merge into a single cluster. Additionally, the noise generated by the model, as well as segmentation faults that extend beyond the edges, can be mistakenly recognized as distinct clusters, further affecting the counting result.

In the course of our investigations, we conducted tests using 3 distinct methods for counting clusters in images. Initially, we implemented the foundational version of the Circular Hough Transform (CHT), a computational technique commonly employed in image processing and computer vision for circle detection. Notably, CHT serves as an extension of the more generalized Hough Transform, with a specific focus on identifying circular shapes [24].

Following a series of tests to determine optimal parameters, we arrived at a refined iteration, setting the minimum radius to 5 and the maximum radius to 60. A sample of the outcomes of these experiments is demonstrated in Figure 3, elucidating the output obtained through the fundamental implementation of the Circular Hough Transform.

It is essential to acknowledge that the algorithm, in its conventional implementation, encounters certain challenges. Specifically, due to the irregular and imprecisely defined circular shapes of wood logs, the algorithm faces difficulties in accurately identifying certain clusters. Furthermore, the current implementation of the algorithm lacks penalization for generating intersections on circles. Consequently, this absence of penalization leads to the identification of multiple circles within the same cluster or circle. It is essential to emphasize that the Circular Hough Transform algorithm is anticipated to yield enhanced results when applied to images constituting the third type of ground truth presented earlier (gradient gray). This expectation arises from the fact that the centroids emphasized by the gray gradient closely approximate a perfect circular shape. Future research endeavors can be directed towards refining this aspect for improved outcomes.

The remaining two algorithms examined in this study involve centroid detection using Morphological Reconstruction and the use of the Connected Components Algorithm for counting the clusters segmented by the model. While these algorithms demonstrated improvements over the Circular Hough Transform (CHT), they encountered a common challenge related to the generation of intersections.

In terms of processing time per image, the Connected Components Algorithm showed an average processing time of 0.58 ms, whereas Morphological Reconstruction averaged 2.38 ms. These results highlight the efficiency of both methods compared with the Circular Hough Transform, which averaged 18.67 ms, but further optimization could reduce computational overhead.

As previously discussed, efforts were made to mitigate the generation of noise and intersections by incorporating mathematical morphology. Although this approach resulted in a partial resolution of the issue, it did not entirely eliminate the problem.

As both solutions achieve similar results in cluster detection, we adopted the classical Connected Components Algorithm due to its computational efficiency and robust performance. This method groups adjacent pixels with the same label into distinct clusters, ensuring precise identification of individual wooden logs. Furthermore, the Connected Components Algorithm demonstrated an average processing time of 0.58 ms per image, outperforming alternative methods like Morphological Reconstruction, which averaged 2.38 ms. This combination of accuracy and efficiency makes it particularly suitable for large-scale or real-time applications.

In this algorithm, the segmented image is represented as a binary matrix $I$, where

$$I_{i,j} = \begin{cases} 1 & \text{if the pixel } (i,j) \text{ belongs to a wooden log,} \\ 0 & \text{otherwise.} \end{cases}$$

The detection of clusters is achieved by grouping all adjacent pixels where $I_{i,j} = 1$ into distinct connected components. To formalize this process, let

1. Set of Object Pixels:

$$P = \{(i,j) \mid I_{i,j} = 1\}, \tag{1}$$

which represents the set of all pixels that belong to wooden logs.

2. Set of Connected Components:

$$C_k = \{(i,j) \mid \text{pixel } (i,j) \text{ belongs to component } k\}, \tag{2}$$

where each connected component $C_k$ contains all pixels in $P$ that are spatially adjacent and belong to the same cluster.

By iterating over the binary matrix $I$, the algorithm efficiently labels each connected component, isolating individual logs and enabling accurate cluster counting.

Regarding the formulation, the connected components labeling algorithm employs a pixel-by-pixel scanning method within the image. This process involves assigning labels to pixels based on observations of neighboring pixels. Subsequently, distinct sets of labels are identified to represent connected components. Finally, the total number of segmented wood logs is determined by counting the number of distinct sets $C_k$.

The flowchart outlining the final process undertaken to obtain the incidence counting results is presented in Figure 8.

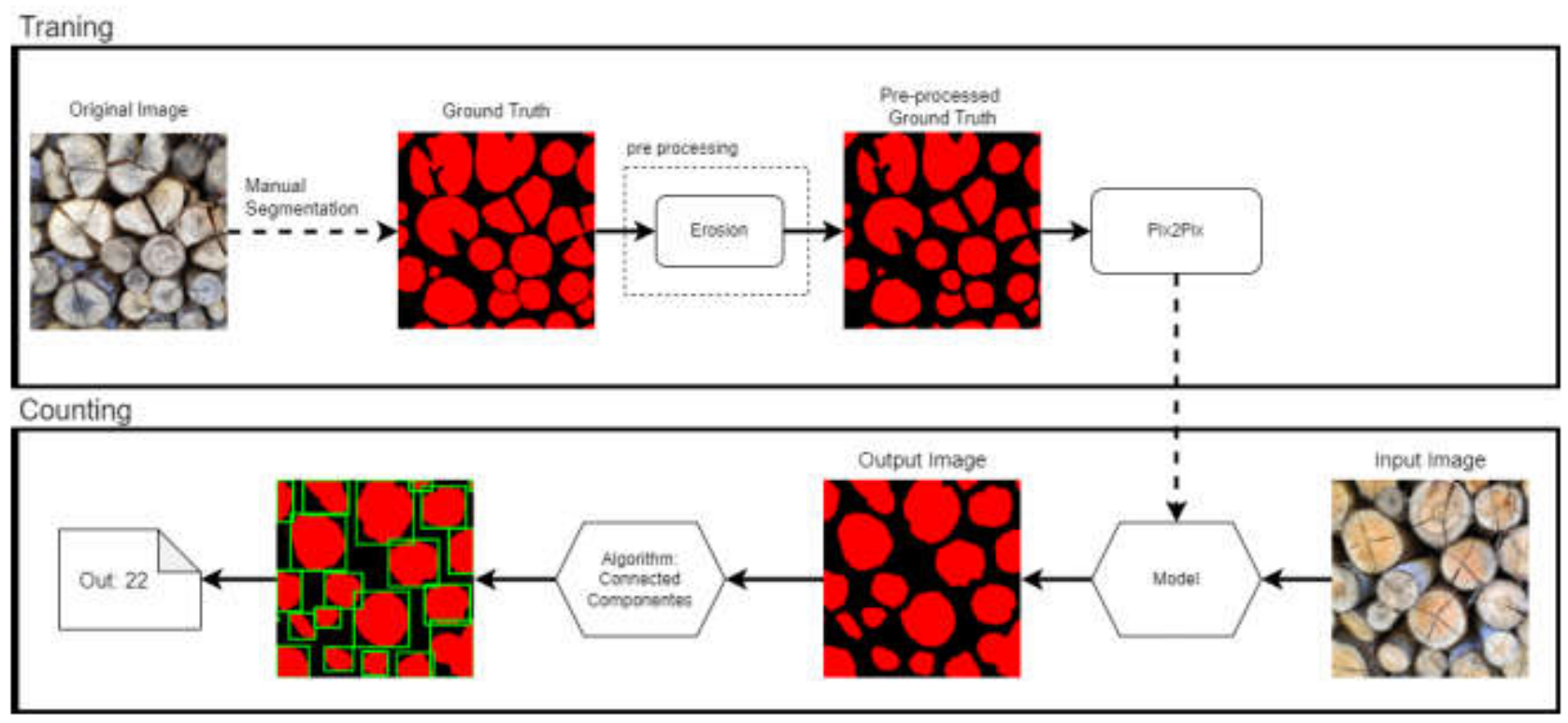

**Figure 8.** Flowchart of the counting method.

Initially, we establish the ground truth over manual segmentation. Subsequently, we subject both the training and test datasets to 15 iterations of erosion. The dataset is then partitioned into 1328 logs (2,970,501 pixels) allocated for training and 448 logs (1,002,097 pixels) for testing purposes.

Following this pre-processing phase, the training dataset undergoes processing within the Pix2Pix framework, incorporating the parameters $input_{nc}$ and $output_{nc}$ to specify the training samples as color-to-color transformations. Pix2Pix generates a model based on this training dataset, which is subsequently employed to process new samples, yielding the resultant segmentation. To ascertain the final counting outcomes, the resulting segmented images are subjected to the Connected Components Algorithm, resulting in the output of the number of logs present in the image.

From the output generated by the connected components method, three observers recorded how many wood logs from the ground truth were represented by the model as part of an intersection and the amount of noise generated by the model compared with the ground truth segmentation.

Following the completion of the model tuning methodology, we proceeded to delineate the statistical scores to be employed for assessing the performance of the model. This evaluation was divided into two main parts: (1) assessing the quality of the model's segmentation and (2) evaluating the methodology's performance for the counting process.

In the first phase, our priority was to determine the quality of the segmentation performed by the Pix2Pix framework. For a systematic analysis, we adopted a pixel-wise evaluation, from which we extracted the following indices: Accuracy, F1 Score (F1), Kappa, and Intersection over Union (IoU).

1. Accuracy: It is the most intuitive form of evaluation. The numerator consists of the sum of all true classifications, whether they are positive or negative, and the denominator consists of the sum of all classifications as follows:

$$Accuracy = \frac{TP + TN}{TP + TN + FP + FN} \tag{3}$$

   where

   - $TP$ (True Positive): number of pixels correctly classified as positive;
   - $TN$ (True Negative): number of pixels correctly classified as negative;
   - $FP$ (False Positive): number of pixels incorrectly classified as positive;
   - $FN$ (False Negative): number of pixels incorrectly classified as negative.

2. F1 Score: Calculated using precision and recall, as follows:

$$F_1 = \frac{2 * precision * recall}{precision + recall} \tag{4}$$

$$precision = \frac{TP}{TP + FP} \tag{5}$$

$$recall = \frac{TP}{TP + FN} \tag{6}$$

3. Kappa: It is a measure commonly used to check the consistency of segmentation, as follows:

$$P_o = \frac{TP + TN}{TP + TN + FP + FN} \tag{7}$$

$$P_e = \frac{(TP + FN) * (TP + FP) + (FP + TN) * (FN + TN)}{(TP + TN + FP + FN)^2} \tag{8}$$

$$Kappa = \frac{P_o - P_e}{1 - P_e} \tag{9}$$

   where

   - $Po$ (proportion of observed agreement) is the ratio of actual agreements to the total number of evaluations;
   - $Pe$ (proportion of expected agreement) is the proportion of agreement that would be expected by chance.

4. Intersection over Union (IoU): It is the ratio between the intersection and union of the ground truth set and the segmentation set to be evaluated. Also known as the Jaccard Index, it measures the overlap between the areas segmented by the model and the actual areas.

$$IoU = \frac{Intersection}{Union} = \frac{TP}{TP + FP + FN} \tag{10}$$

These indices provide a comprehensive score for assessing different aspects of the model's segmentation quality, contributing to a more complete evaluation of performance and comparisons for future works.

After the index definition is used to evaluate segmentation performance, we adopt the subsequent formula to appraise the counting methodology. We designate this formula as the Intersection-Sensitive Score (ISS), and it is expressed as

$$ISS = \frac{CI}{CI + E + I + N} \tag{11}$$

where

- $CI$ represents the number of logs that are correctly identified (True Positive);
- $E$ represents the number of non-identified logs (False Negative);
- $I$ represents the number of logs classified as intersections;
- $N$ represents the number of noises classified as logs (False Positive).

To prevent any misinterpretation, it is crucial to provide a mathematical explanation of how $I$ is defined: Let $A$, $B$, and $C$ represent clusters individually depicting the segmentation of a wood_log. When referring to the intersection of $A$ and $B$ denoted as $A \cap B$, then $I = 2$, it means that there are two different logs classified as one single cluster.

Similarly, when considering the intersection of $A$, $B$, and $C$, denoted as $A \cap B \cap C$, then $I = 3$, it means that we have three different wood logs being classified as one single cluster. This pattern continues for additional intersections. The visual representation of the idea mentioned above can be seen in Figure 9.

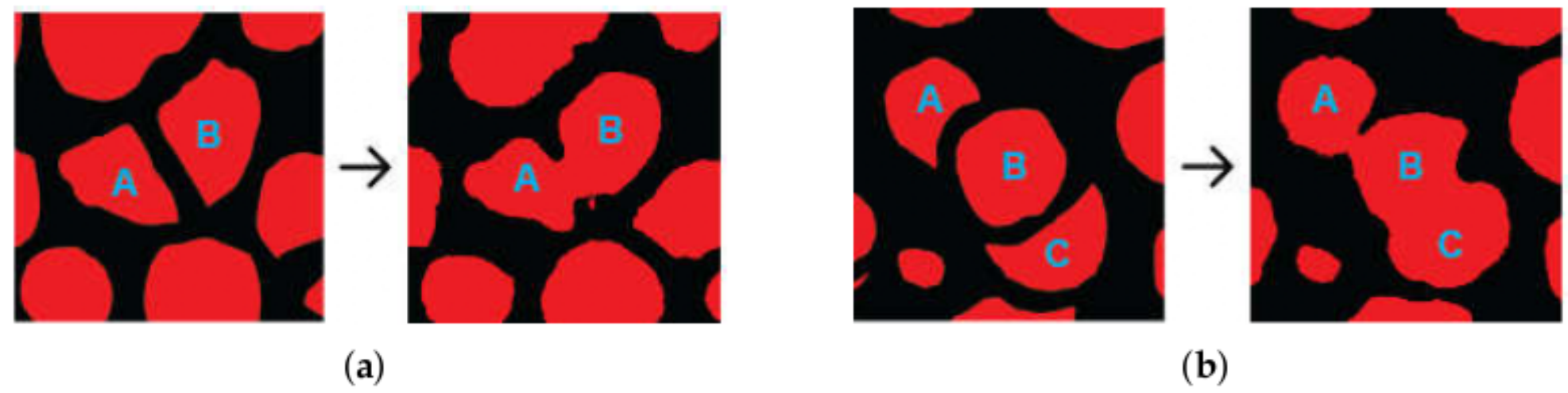

**Figure 9.** Visual representation with examples of variable $I$ definition: (**a**) example: $I = 2$; (**b**) example: $I = 3$.

We introduce this nomenclature to better assess the model's performance, specifically when other measurements do not account for intersection when generating a score. Additionally, to clearly distinguish between discussions of accuracy in segmentation and performance in log counting, whenever Accuracy pixel is mentioned, it pertains to segmentation, and whenever Accuracy logs is referred to, it pertains to accuracy as mentioned in Equation (3) but is applied to log counting.

As a potential practical application of the methodology proposed in this study, we present a workflow that demonstrates how the model could be integrated into forestry operations. This application highlights the feasibility of using the proposed approach to streamline log segmentation and counting in real-world scenarios, from image capture in the field to reporting and logistics integration. This aligns with current discussions in the literature of integrating AI tools into forestry operations to improve accuracy and efficiency [25,26]. Figure 10 provides an overview of this workflow, and the main stages are detailed below.

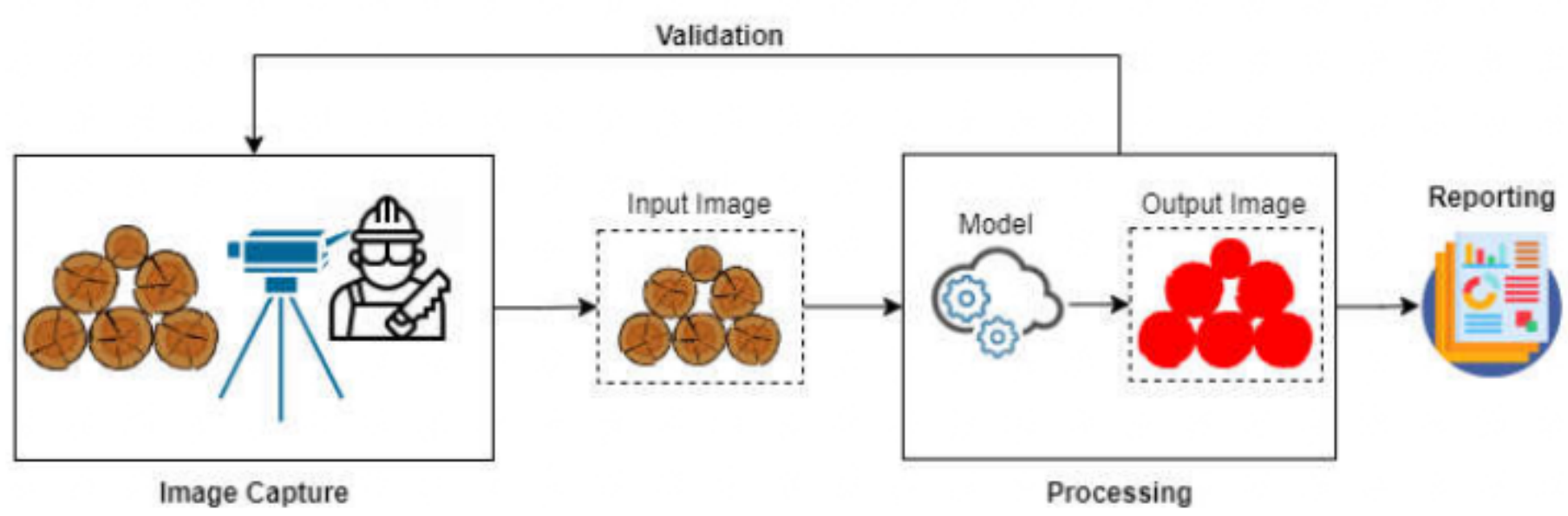

**Figure 10.** Automated log counting workflow for forestry operations.

The workflow begins with image capture, where a mobile device equipped with the proposed model is used to photograph log piles directly in the field. The application guides the operator to properly frame the logs, ensuring high-quality images for subsequent processing. This step eliminates the need for additional equipment, such as drones or specialized cameras, reducing costs and simplifying field operations.

Next, the captured images undergo processing on cloud. The model pre-processes the images (e.g., noise removal and brightness adjustments) before segmenting and counting the logs. The results, including highlighted and numbered logs, are displayed instantly on the application, allowing operators to review the output immediately.

In the validation stage, operators can refine the results using the application's interactive tools. This allows them to add or remove logs identified incorrectly, ensuring high accuracy and adapting the system to varying field conditions. This combination of automation and manual refinement makes the method both efficient and flexible.

The validated data are then handled in the storage and transfer stage, where they are stored locally on the device and, when connectivity is available, uploaded to a central server or logistics system. For remote operations without internet access, the data remain on the device until synchronization becomes possible, ensuring uninterrupted workflow continuity.

Finally, in the reporting stage, the processed data are used to generate automated reports. These reports include key scores such as the total number of logs, estimated volume, and identified anomalies, which can be directly shared with stakeholders or integrated into logistical planning systems to optimize resource allocation and transportation.

## 3. Results

This section will be bifurcated in two parts. Initially, we address an in-depth experimentation with the segmentation process employing the Pix2Pix framework within the context of the scenarios that we formulated. Subsequently, we present the data resulting from the tests carried out to evaluate the counting methodology proposed. It is important to emphasize that the model developed during the segmentation phase does not employ the same parameters used in training the model designed for the optimized generation of labeled images for counting.

### *3.1. Segmentation*

The results demonstrate that for the complex task of wood log segmentation, which involves handling variations in lighting, overlapping objects, and diverse spatial configurations, Pix2Pix exhibits superior adaptability and generalization compared with other methods. Despite its significantly larger number of parameters, Pix2Pix effectively leverages its adversarial framework and architectural design to achieve consistent performance, making it a robust choice for such challenging scenarios. In contrast, simplified models, such as the Vanilla Autoencoder and the Vanilla U-Net, were specifically modified in this study to reduce their number of parameters and mitigate overfitting when applied to limited datasets. The Autoencoder was simplified to approximately 20,000 parameters, while the U-Net was reduced to 15,000 parameters. Although these modifications successfully minimized overfitting, as evident from their validation loss curves, the simplified models struggled to generalize effectively under the complex conditions required for wood log segmentation. A detailed comparison of the model parameters, validation loss, segmentation accuracy, and generalization capabilities is presented in Table 1.

**Table 1.** Comparison of model performance and generalization capabilities.

| Model | Parameters | Validation Loss (MSE) | Segmentation Accuracy (%) | Generalization Notes |
|---|---|---|---|---|
| Vanilla Autoencoder [27] | ~20,000 | 0.062 | 85.4 | Limited generalization |
| Vanilla U-Net [28] | ~15,000 | 0.059 | 86.9 | Limited generalization |
| Pix2Pix (Proposed) | 57,183,620 | 0.032 | 94.1 | Superior generalization |

Figure 11 provides a comparative graph of training and validation losses for the three models. The overfitting behavior in the original architectures was mitigated in the modified Vanilla Autoencoder and Vanilla U-Net due to the reduction in parameters. However, this simplification came at the cost of reduced capacity to generalize to the complex patterns required for this task. In contrast, Pix2Pix demonstrates consistent performance without signs of overfitting, even with its significantly larger number of parameters.

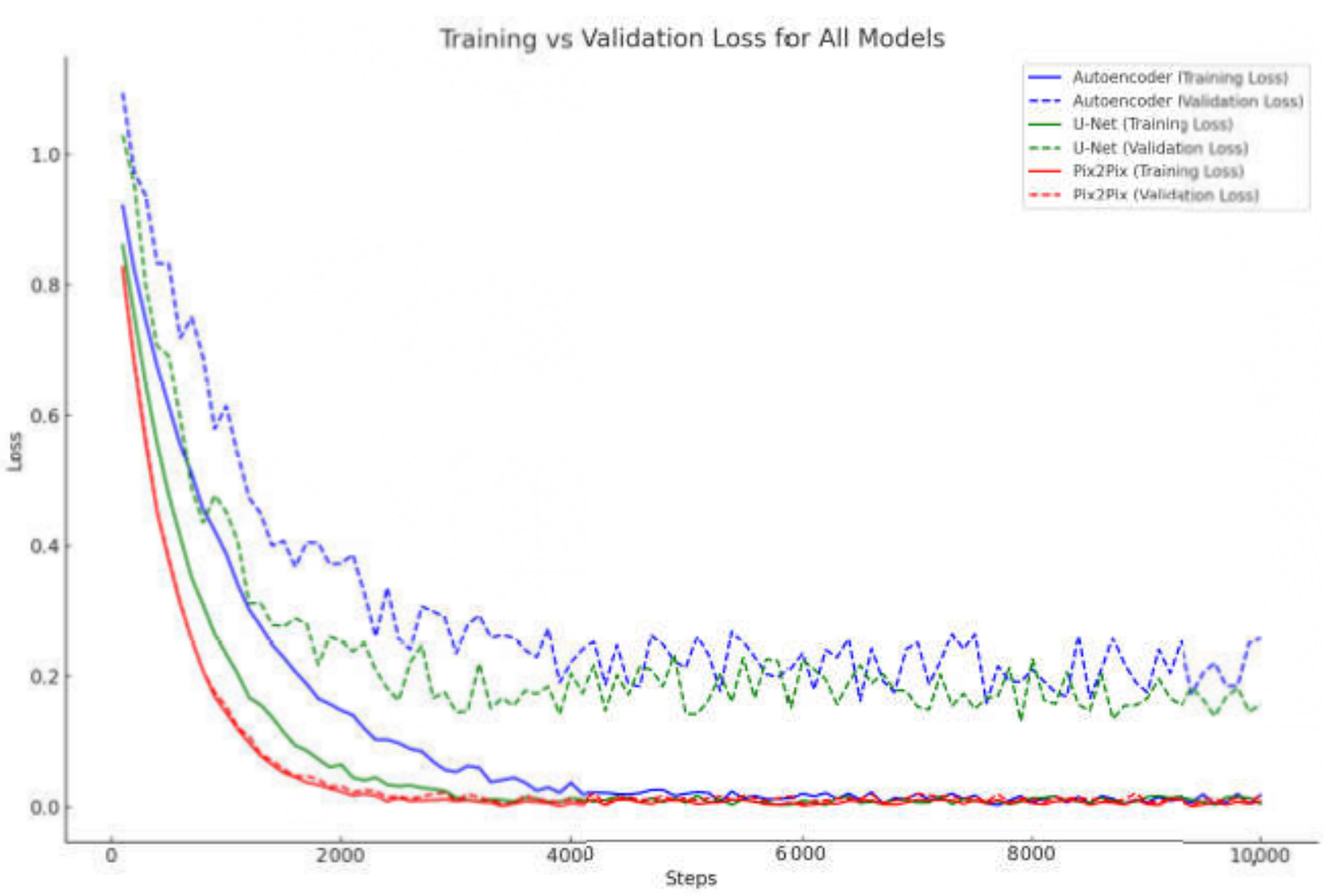

**Figure 11.** Comparison of training and validation losses for modified Vanilla Autoencoder, modified Vanilla U-Net, and Pix2Pix.

To further investigate qualitative differences, Figure 12 presents segmentation outputs generated by each architecture. The Pix2Pix model shows clear superiority, effectively capturing log details and significantly reducing noise and intersection generation compared with the simpler models, including the modified Autoencoder and U-Net.

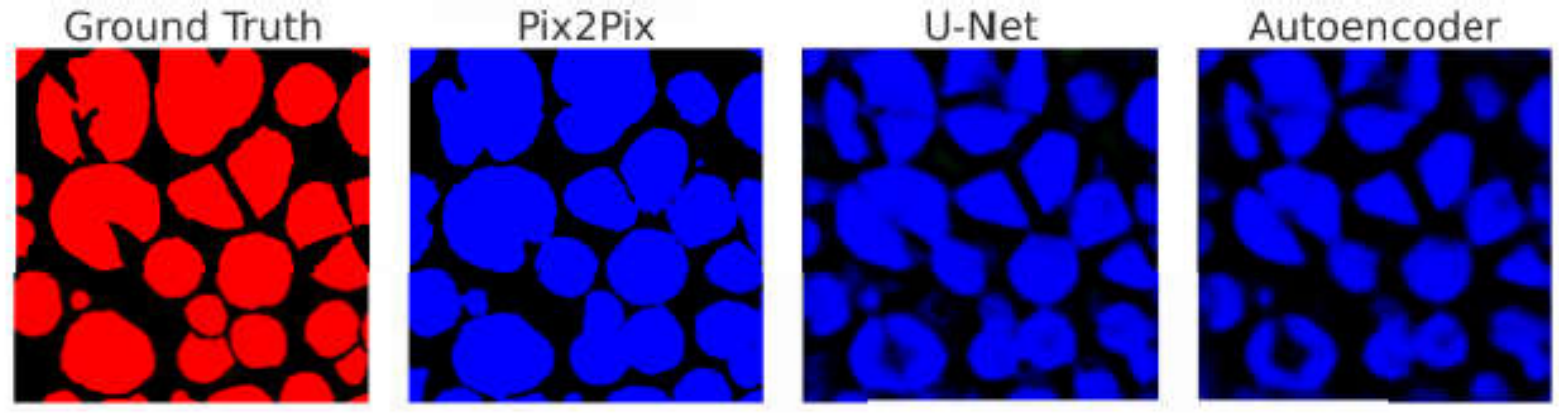

**Figure 12.** Qualitative comparison of segmentations: modified Vanilla Autoencoder, modified Vanilla U-Net, Pix2Pix, and ground truth.

The experimental analysis highlights the importance of aligning model complexity with dataset size and diversity. While the modified Vanilla Autoencoder and U-Net were effective in reducing overfitting due to their lightweight design, they lacked the capacity to learn the complex patterns required for wood log segmentation. In contrast, Pix2Pix fully leveraged the dataset's variability, offering superior and consistent performance in segmentation and log counting.

Moreover, the dataset expansion proved pivotal in enhancing model performance, both quantitatively and qualitatively. The broader and more diverse dataset enabled the Pix2Pix model to better generalize across varying conditions, improving its ability to identify precise log boundaries and to handle densely packed regions. This result is consistent with findings from [29], which demonstrated that increasing the size and granularity of training datasets significantly enhances the robustness and fidelity of segmentation models, particularly those leveraging generative adversarial networks. These insights underscore the critical role of dataset diversity in achieving high-quality segmentation and counting outcomes.

Applying the methodology illustrated in Figure 2, we designed the experimental setup to evaluate segmentation quality against the ground truth. The steps, from model training to segmentation quality assessment, are detailed in the flowchart.

The dataset consists of images manually segmented into wood logs. From this, 1328 logs (2,970,501 pixels) were allocated for training and 448 logs (1,002,097 pixels) for testing. The test images were fed into the trained models, generating segmentations that were compared pixel by pixel against the ground truth using a quality evaluation algorithm. Performance indices such as pixel-wise accuracy, F1 score, Cohen's Kappa, and IoU were extracted to assess model effectiveness.

A subset of the test results is presented in Table 2, and visual comparisons are shown in Figure 13. These results further emphasize Pix2Pix's ability to maintain high segmentation accuracy while effectively generalizing across diverse input conditions.

**Table 2.** Sample of segmentation results presenting indexes: $Accuracy_{pixel}$, F1, Kappa, and IoU.

| Image Index | $Accuracy_{pixel}$ | F1 Score | Kappa | IoU |
|---|---|---|---|---|
| 71 | 0.962 | 0.907 | 0.884 | 0.831 |
| 17 | 0.956 | 0.897 | 0.869 | 0.814 |
| 26 | 0.970 | 0.933 | 0.914 | 0.875 |
| 34 | 0.961 | 0.910 | 0.885 | 0.835 |
| 59 | 0.968 | 0.925 | 0.905 | 0.861 |
| 12 | 0.971 | 0.932 | 0.913 | 0.872 |
| 55 | 0.957 | 0.879 | 0.853 | 0.784 |
| 10 | 0.970 | 0.924 | 0.905 | 0.858 |
| 2 | 0.969 | 0.917 | 0.898 | 0.847 |
| 46 | 0.964 | 0.908 | 0.886 | 0.831 |

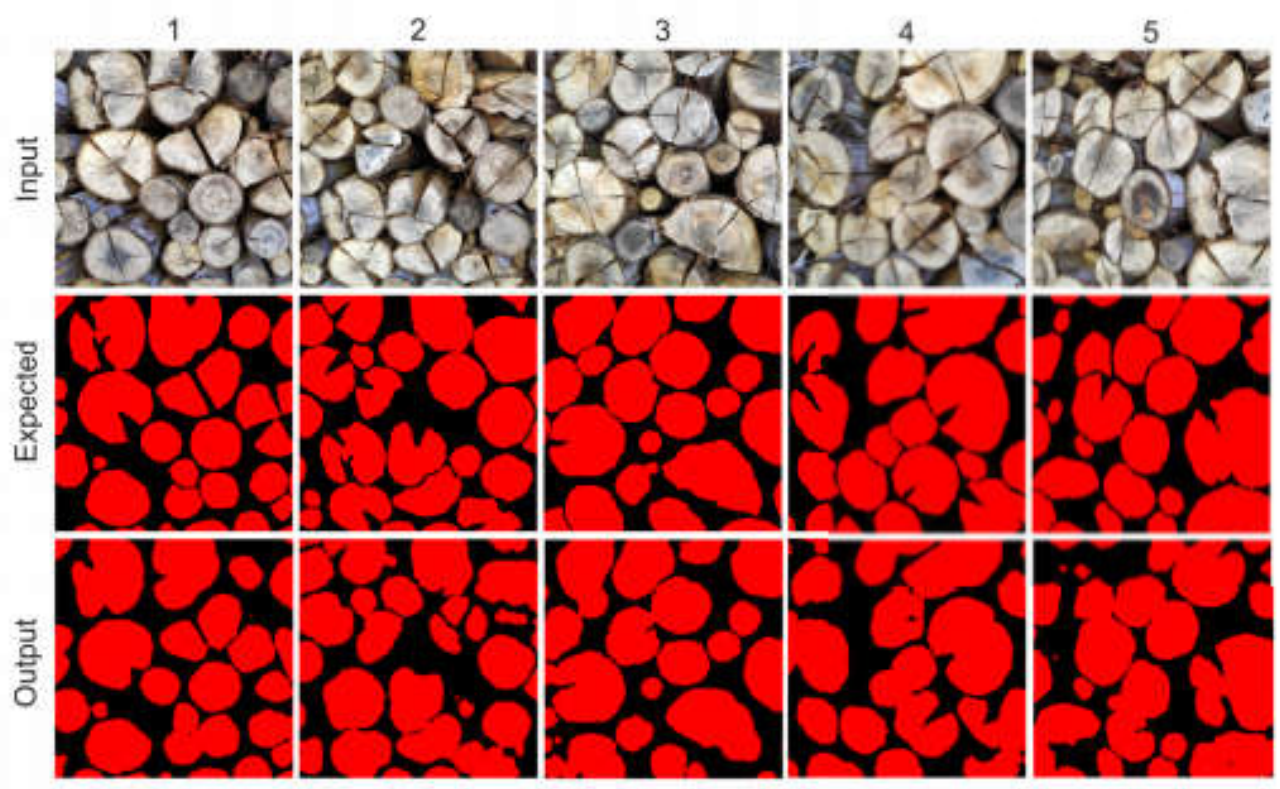

**Figure 13.** Segmentation results sample.

Analyzing the results presented in Table 2 together with Figure 13, the complexity of visual interpretation versus numerical precision in the segmentation task becomes evident. While an average $\text{Accuracy}_{pixel}$ of 96.4% might initially imply underperformance.

Upon deepening the analysis and examining specific details, such as the second sample, it becomes clear that interpreting the model in relation to subtle elements, such as cracks in the wood logs, can result in notable discrepancies between visual expectation and quantitative assessments.

It is observed that during the ground truth segmentation, cracks in wood logs were ignored and segmented as pixels that are not part of a wood_log, whereas the model interpreted these cracks as a single piece, incorporating them into the wood_log. Conversely, inverse situations also occurred, where certain features segmented by the model were overlooked or treated differently in the ground truth, leading to discrepancies in the quantitative interpretation of the segmentation results. However, visually, the model executed the segmentation accurately.

### *3.2. Counting*

During this experimental phase, our main objective was to achieve a segmentation that was later used to count the number of incidences of logs. To accomplish this, we employed an iterative process, subjecting the ground truth to 15 consecutive erosion operations before embarking on the model training phase. The sequential steps of this process are illustrated in Figure 8.

In both the training and testing datasets, images underwent manual segmentation and were subjected to 15 iterations of erosion as a pre-processing step. Specifically, 1328 logs (2,970,501 pixels) were allocated for training and 448 logs (1,002,097 pixels) for testing.

Following the training phase, test images were processed through the model, generating segmentations that underwent evaluation using the Connected Components Algorithm. The evaluation algorithm, benchmarked against the ground truth, extracted the number of logs identified.

Subsequently, three observers determined the count of correctly identified logs, the number of logs involved in intersections, and the instances of noise generated by the model. Ultimately, performance indices, encompassing $\text{Accuracy}_{\text{logs}}$ and the Intersection Sensitive Score (ISS), were calculated.

A concise summary of the results is presented in Table 3, complemented by visual representations in Figure 14.

**Table 3.** Random sample of log counting results.

| Image Index | Expected Number of Logs | Output | Correctly Identified (CI) | Intersecting Logs (I) | Noise (N) | ISS (%) |
|---|---|---|---|---|---|---|
| 71 | 31 | 29 | 22 | 7 | 2 | 70.97 |
| 17 | 26 | 23 | 13 | 10 | 3 | 50.00 |
| 26 | 22 | 23 | 13 | 5 | 4 | 56.52 |
| 34 | 19 | 19 | 11 | 6 | 1 | 61.11 |
| 59 | 26 | 24 | 16 | 7 | 3 | 61.54 |
| 12 | 23 | 22 | 17 | 4 | 2 | 68.00 |
| 55 | 28 | 29 | 10 | 12 | 5 | 37.04 |
| 10 | 23 | 23 | 12 | 8 | 2 | 48.00 |
| 2 | 21 | 23 | 16 | 3 | 2 | 76.19 |
| 46 | 21 | 19 | 14 | 6 | 1 | 66.67 |

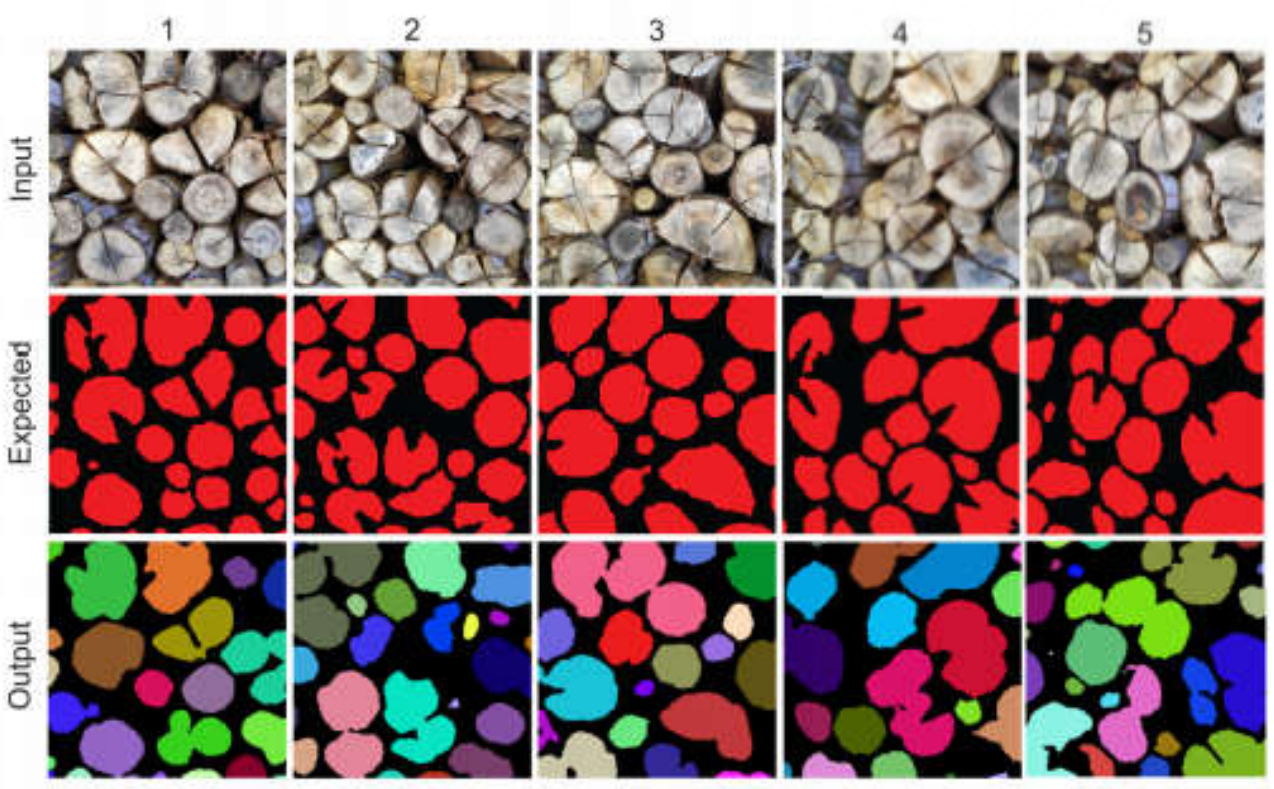

**Figure 14.** Visual representation of random samples in log segmentation and counting.

Similar to the segmentation results, it is essential to highlight that an $Accuracy_{logs}$ exceeding 92.3% does not adequately define the precision of counting. A thorough visual examination of the images is necessary for a comprehensive evaluation. As a solution, the Intersection Sensitive Score (ISS) is introduced, enabling the assessment of both generated intersections and the presence of noise in the final results.

It is pertinent to acknowledge that the acquired data are intricately linked to the ground truth, and as previously highlighted, the classification process is notably affected by human bias. In concordance with the observations derived from the segmentation results, the observer retains the discretion to selectively segment solely the countable faces of logs, as perceived from their standpoint in the image. The subjective interpretation of smaller logs or those with partially visible faces is contingent upon the observer assessment.

On the other hand, the model tends to segment any pixel that could be part of a heartwood, regardless of whether it is entirely within the frame or not. Consequently, many of the generated noises are interpretative, as the model chooses to segment a cluster not considered by the observer in the ground truth, but it is part of a fragment of a log that is not entirely visible in the image. This leads to a discrepancy in the counting of logs, intersections generated, and mainly noise generation.

In Figure 15, it is noteworthy that in both instances, the model generated justifiable noise. In Figure 15a, on the left side, a minor section of a wood_log remains visible within the image frame. This wood_log was deliberately omitted during the ground truth segmentation by the observer, while the model identified this small portion as part of the wood_log, resulting in noise when compared with the ground truth; however, it is indeed a wood_log in the input image. Similarly, in Figure 15b, at the top, the model identifies two small clusters as part of a wood_log. The smaller cluster, situated closer to the center, represents a segmentable fraction of a log, even though it is part of a small branch. The second noise at the top of the same image mirrors the situation in Figure 15a, where the model segmented a small portion of a wood_log at the edge of the input image.

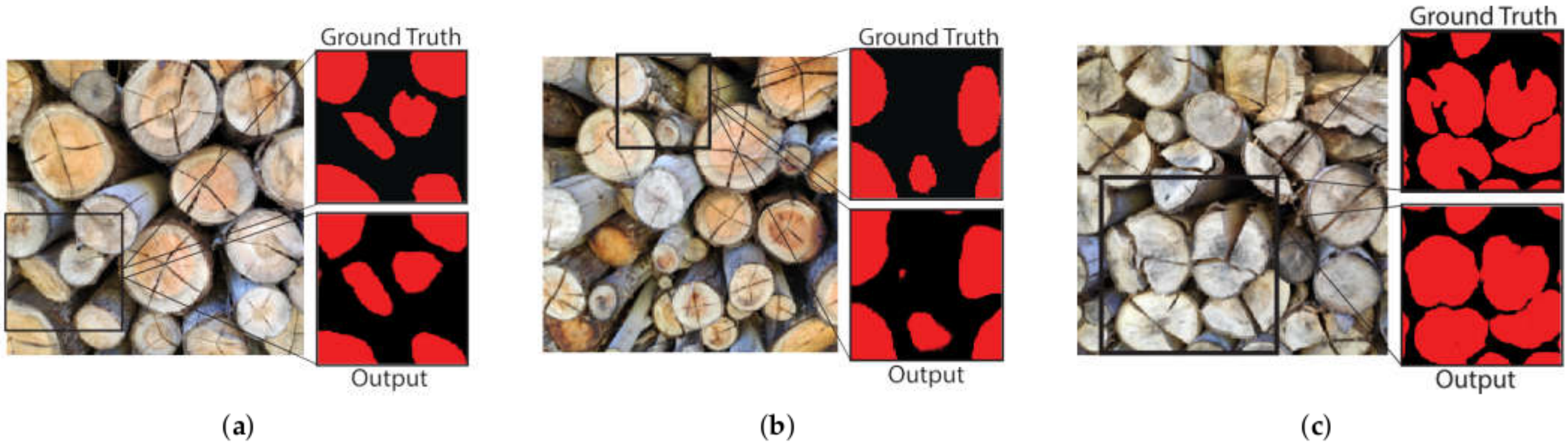

(a) (b) (c)

**Figure 15.** Closer view and analysis for sample image indices 12, 55, and 17: (**a**) lateral segmentation noise observed in index 12, (**b**) top segmentation noise observed in index 55, and (**c**) crack segmentation discrepancies observed in index 17. Each subfigure presents the ground truth and output segmentation results for a detailed comparison.

It is worth noticing that the model is still generating intersections and a few noises. As mentioned above, intersections commonly occur, particularly in cases of occlusion where one wood_log overlaps another, while noises are predominantly observed along the edges of the input image. With the challenge of generating intersections and noise persists, employing a series of dynamic erosions with Euclidean distance instead of a fixed kernel in a few iterations might prove effective in mitigating most intersections but should not be the standard solution. Furthermore, a post-processing step on the image generated by the model, which implements some constraint solutions, may be sufficient to eliminate most of the partial wood logs (noises) that were segmented. This would result in preserving only wood logs that are entirely within the frame or that meet a threshold of at least 60 pixels, for example.

However, these cases highlight broader challenges in addressing edge effects and noise generation in automated segmentation models. For instance, edge effects occur when objects near the boundaries of the image are partially visible, leading to segmentation inconsistencies. This is particularly problematic in Figure 15a, where the model's sensitivity to small portions of logs at the edges results in outputs that diverge from human annotations. Such discrepancies suggest the need for a more robust edge-handling strategy, such as integrating padding techniques during pre-processing or implementing post-processing rules to discount small segments near the edges.

Furthermore, residual intersection issues remain a challenge. These issues are exemplified in Figure 15b, where overlapping branches and logs result in misclassification. Future work could explore enhanced intersection-specific loss functions or graph-based post-processing techniques to better delineate overlapping objects.

Additionally, the inclusion of a more nuanced noise evaluation framework could strengthen the analysis. For example, categorizing noise into "justifiable" (e.g., segments that align with portions of logs) and "non-justifiable" (e.g., misclassified artifacts) would provide a deeper insight into the model's performance. This could also guide iterative improvements in both training data annotation and model architecture design.

## 4. Discussion

When critically evaluating related works, it is noteworthy that the mentioned authors predominantly concentrate on accuracy and do not furnish additional essential indices for a comprehensive comparison, with the exception of [10]. Our attempts to secure access to the utilized datasets, unfortunately, did not yield successful outcomes. However, even if other indices or scores were available, the fairness of a comparison between the presented works would still be compromised due to the use of distinct datasets. Therefore, conducting a more detailed analysis for comparison may not be feasible. Nevertheless, it is crucial to highlight that our study has been meticulously documented, enabling any collaborative or similar approach to easily utilize the same dataset and indices obtained during development.

In order to explore new approaches in model training, we conducted an interesting experiment, as illustrated in Figure 16. In this scenario, we reversed the input and expected output during training, using a training set composed of 15 images. The results obtained revealed an intriguing perspective on the adaptability of the Pix2Pix model.

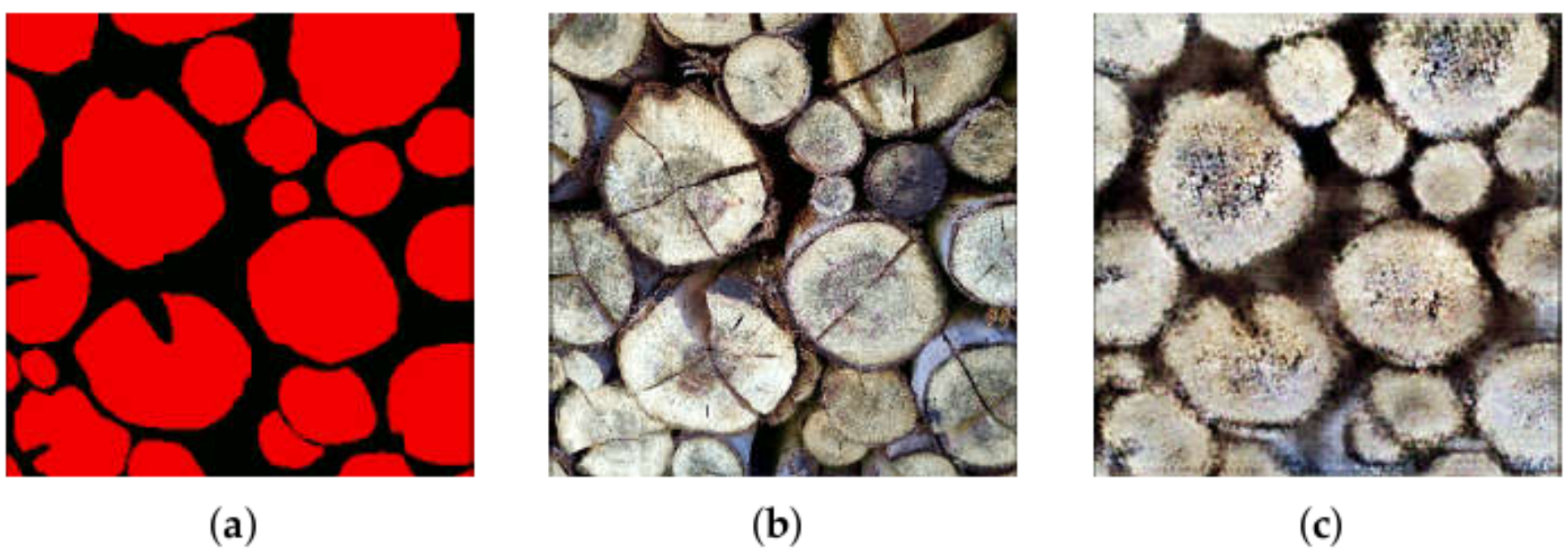

(a) (b) (c)

**Figure 16.** Input and expected output inversion experiment in Pix2Pix training. This experiment demonstrates the adaptability of the Pix2Pix model when reversing input and expected outputs during training using a dataset of 15 images. The figure shows (**a**) input segmentation masks, (**b**) expected wood log textures, and (**c**) output generated by the model. The results highlight the potential of Pix2Pix in generating realistic representations and its application in procedurally creating visual content for natural elements.

Remarkably, the generated representations exhibit notable similarities to real wood logs. This finding suggests the possibility of further enhancing fidelity through parameter adjustments and more refined training. These preliminary results indicate a promising application of Pix2Pix in procedurally generating wood logs, providing an innovative approach to visual content creation. This establishes a foundation for more in-depth investigations into the use of Pix2Pix in the procedural generation of natural elements.

Regarding the ISS measure, introduced as a contribution for evaluating the models' performance, it is worth noting that while it was developed specifically for the context of wood log segmentation, it holds potential for broader applicability in studies involving object segmentation as part of the instance identification process (instance segmentation). ISS adds value by incorporating the intersection variable as a critical factor in determining the score. However, its implementation requires caution due to the subjectivity involved, as no automated method currently exists to detect intersections. This process must be carried out manually, demanding a clear definition of what constitutes an intersection, as we mentioned in this document in Figure 9. Furthermore, a detailed discussion on this criterion is recommended to ensure the replicability of results.

It is worth mentioning that the results of this study demonstrate the effectiveness of the proposed method for log counting under controlled conditions, providing a strong foundation for its applicability. However, one critical aspect that remains unexplored is the comparison with manual counting under real-world industrial settings. Although we performed a manual count with three independent observers to validate the model's performance, this approach was limited to controlled environments and does not replicate the practical conditions in which industrial workers conduct manual counts. In real-world scenarios, various factors—such as worker fatigue, environmental conditions (e.g., lighting, weather), and operational time constraints—can influence the accuracy and reliability of manual counting. These factors introduce complexities that are absent in controlled validations, making it essential to investigate the performance of the proposed method directly in such settings. A comparative evaluation between the proposed method and manual counting in the field would not only provide insights into its practical advantages but also highlight areas where further improvements may be required.

Additionally, a real-world case study would allow the proposed methodology to be tested for robustness and usability in operational environments. Such a study could compare the time required, accuracy, and overall efficiency of manual counting with the automated approach, offering tangible evidence of the practical benefits of the proposed method. This would be particularly valuable in scenarios where resources for advanced technologies like LiDAR are unavailable. While manual counting is the standard approach in many small-to-medium-scale industries, its inherent limitations—such as subjectivity, susceptibility to human error, and labor intensity—can significantly affect productivity. The proposed method, leveraging RGB cameras and 2D image processing, presents itself as an accessible and scalable alternative that could bridge the gap between high-end LiDAR solutions and traditional manual approaches. Incorporating these comparisons into future work would strengthen the argument for adopting automated methods in resource-constrained industries.

Furthermore, while LiDAR-based methods are renowned for their precision and accuracy, their reliance on specialized and costly hardware limits their applicability for smaller operations. These methods require calibrated laser scanners and controlled conditions for data acquisition, both of which present significant barriers to widespread adoption. In contrast, the proposed method solely relies on RGB cameras, making it far more accessible for less capital-intensive industries. However, a direct comparison with LiDAR methods in specific scenarios, such as large-scale log counting operations, could further validate the proposed approach's effectiveness and versatility. Future work should aim to address these limitations by conducting a comprehensive field study that includes comparisons with manual counting and, where feasible, LiDAR-based approaches. This would provide a more holistic evaluation of the methodology's strengths and weaknesses, paving the way for its adoption across a wider range of applications.

The potentially superior performance of Pix2Pix is influenced by the dataset's characteristics, as it relies on supervised learning. Generalization to other datasets depends on how well their distributions align with the training data. For significantly different datasets, fine-tuning or transfer learning may be necessary to maintain performance. However, in general, Pix2Pix has demonstrated robust performance across various domains [30,31].

## 5. Future Research

Numerous alternative approaches exist for the counting stage, and we conducted experiments using Circular Hough Transform. This segmentation technique has the potential to yield good results, particularly in isolated cases and when applied to close-range images. Figure 17 presents preliminary tests carried out where the reference ground truth passed to the model consists of the generation of clusters in gray scale, where the center presents absolute white and the edges lose color intensity.

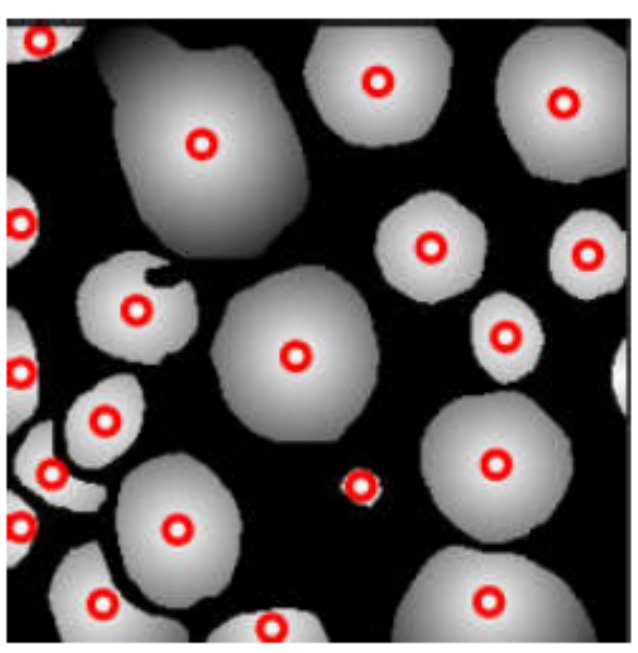

**Figure 17.** Circular Hough Transform applied to gray circular gradient Morphological Reconstruction.

During the tests, we determined a circle of fixed size in the implementation of an algorithm respecting the Circular Hough Transform paradigm, combining this with a database where the centroids are highlighted in a circular shape where there is greater density in the segmentation, counting through the incidence of centroids, which can be an effective approach. With this approach, the main problem of Circular Hough Transform in the task of counting irregular objects would be solved. Another point to be highlighted is the fact that by having a fixed centroid size, the size of the wood_log is irrelevant for it to be classified and correctly accounted for.

It is worth mentioning that, with a fixed centroid size, it is possible to prepare the Circular Hough Transform algorithm with the same radius size so that it correctly identifies each cluster. It is important to emphasize that the absence of intersections is essential if the gray gradient is applied to each cluster, unlike pixel density mapping.

Further, we can explore the application of wooden wood_log face segmentation for enhanced pile volume calculation. Although it is not one of the main objectives of the present work, it is important to highlight that the segmentation of the wood logs can be a useful artifact to calculate the approximate volume of a pile where the logs are perfectly rounded. There are two main ways to calculate the volume of a pile: The first is to calculate the volume from a pile in a row, that is, a pile organized in order to facilitate loading. The second approach, although more laborious, especially when carried out without technological assistance, offers a more accurate volume result. This consists of calculating the volume of each log individually, as follows:

$$Lv = \pi r^2 c \tag{12}$$

where $Lv$ is the log volume, $r$ the radius of one of the faces, and $c$ the log depth.

Therefore, the calculation of the volume of a non-rowed pile can be obtained using the following formula:

$$V = \sum_{n=0}^{T} \pi r_n^2 c_n \tag{13}$$

where $V$ is the total volume of the pile, $T$ is the number of logs that make up the pile, $d$ is the diameter of the log face, and $c$ is the depth of the log.

Shifting the focus to the database used for model training, it is noteworthy that the proposed methodology demonstrates versatility when considering the subject of study. In other words, if there is a segmented database of other objects or individuals, counting them using the same methodological approach is feasible.

An important avenue for future work involves exploring the operational feasibility and cost implications of integrating the proposed method into existing forestry workflows. While this study demonstrates the method's potential for accurate and efficient log counting, further investigation is required to evaluate its real-world applicability and economic impact compared with traditional and complex approaches such as manual counting and LiDAR-based methods.

The proposed methodology offers a compelling balance between cost and efficiency. Unlike LiDAR-based approaches, which require expensive hardware such as laser scanners and precise calibration protocols, this method solely relies on RGB cameras and computational tools, making it a significantly more affordable alternative. This lower cost of implementation and maintenance positions the proposed method as an attractive option for small-to-medium-scale forestry operations, where an advanced LiDAR infrastructure may not be feasible. Additionally, the method's automation capabilities provide a distinct advantage over manual counting by reducing labor requirements and minimizing human errors, which are common in traditional workflows, particularly in challenging operational environments.

Future studies could focus on conducting comprehensive field trials to evaluate the scalability and usability of the proposed method in real-world forestry operations. These trials could compare the time, cost, and accuracy of the proposed approach with both manual and LiDAR-based methods, offering a more holistic perspective on its operational feasibility. Moreover, the development of training protocols and support materials for forestry personnel would be an essential step to facilitate the adoption of this technology. Such efforts would provide critical insights into the long-term economic and operational benefits of the proposed methodology, further validating its role as a practical and cost-effective solution for log counting in diverse forestry scenarios.

## 6. Conclusions

We introduced a novel approach for wood log counting, leveraging Conditional Generative Adversarial Networks (cGANs) combined with advanced image processing techniques. As part of this study, we proposed the Intersection Sensitive Score (ISS), a score designed to evaluate the model's performance by accounting for intersection challenges inherent in object segmentation tasks. Our approach for wood log counting demonstrated robust performance. The pixel-level accuracy achieved an average of 96.4%, indicating highly precise segmentation, while the log counting accuracy reached 92.3%, demonstrating the effectiveness of our method in object quantification. The F1 scores, ranging from 0.879 to 0.933, reflect a strong balance between precision and recall. The Kappa coefficient, measuring the agreement between predicted and actual segmentation, showed excellent results between 0.853 and 0.914, while Intersection over Union (IoU) values of 0.784 to 0.875 demonstrated an effective overlap between predicted and ground truth segmentations. The proposed Intersection Sensitive Score (ISS) provided additional insight into the model's performance in handling intersecting logs, with values ranging from 37.04% to 76.19%, highlighting areas where the model excels and where there is room for improvement in handling complex log arrangements.

From an implementation perspective, the solution demonstrated efficient processing capabilities with an average inference time of 0.713 s per image when running on an NVIDIA T4 GPU. This performance level makes it suitable for deployment as a cloud-based system using a microservice architecture, where clients could submit images through a RESTful API interface and receive segmentation results along with log counts in near real time. This architectural approach allows for flexible deployment options and integration with existing industrial systems, making it particularly suitable for real-world applications in the forestry sector.

Looking ahead, as computer vision and machine learning technologies continue to evolve, we anticipate that the integration of cGANs with complementary image processing techniques will play a pivotal role in driving innovation in object counting tasks. The methodology demonstrates significant potential to advance object counting across various real-world applications, setting itself apart through the innovative application of cGANs. Beyond improving stock counting in industrial settings and enabling precise diagnostics in laboratory environments, this approach shows promise in the construction sector, where accurate wood log measurements can facilitate material estimation and minimize resource waste. By delivering precise and reliable data, the methodology contributes to enhancing operational efficiency and promoting sustainability across forestry and related industries.

For future research, we explored the application of the Circular Hough Transform as an alternative to the Connected Components Algorithm, particularly to mitigate limitations such as intersections and noise in dense object counting scenarios. Moreover, we proposed a potential extension of the methodology to estimate the volume of wood log piles using the segmentation outputs generated by the model. These future directions aim to enhance the practical utility and scope of the proposed method while maintaining the high accuracy standards demonstrated in our current results.

**Author Contributions:** Conceptualization, J.V.C.M.; methodology, J.V.C.M., G.T. and E.M.F.D.; software, G.T. (deep learning model development and training), E.M.F.D. (image processing implementation), and É.O.R. (model optimization); validation, G.T., G.B.V. (model validation and parameter analysis) and G.A.O. (dataset validation); formal analysis, M.T. (statistical analysis); investigation, J.V.C.M., G.A.O. and E.O.R.; resources, J.V.C.M.; data curation, G.A.O. (dataset collection and curation); writing—original draft preparation, J.V.C.M. (main manuscript writing); writing—review and editing, G.B.V. (manuscript refinement and review responses); visualization, M.T. (metrics interpretation and visual representations); supervision, J.V.C.M. (project leadership); project administration, J.V.C.M. (project coordination). All authors have read and agreed to the published version of the manuscript.

**Funding:** This research received no external funding. However, the authors are deeply grateful to the Universidade Federal de Itajubá for its support in covering the APC.

**Data Availability Statement:** The data presented in this study and the trained model weights are openly available in our GitHub repository at https://github.com/JoviMazzochin/tcc-utfpr (accessed on 8 January 2025). The dataset is regularly updated as new data become available, and the pre-trained weights can be used for transfer learning applications. Additional data or information about this research can be requested from the corresponding author.

**Conflicts of Interest:** The authors declare no conflicts of interest.